%% file: main.tex
\documentclass{article}

\PassOptionsToPackage{numbers, compress}{natbib}
\usepackage[preprint]{neurips_2026}

\usepackage[utf8]{inputenc} 
\usepackage[T1]{fontenc}    
\usepackage{hyperref}       
\usepackage{url}            
\usepackage{booktabs}       
\usepackage{amsfonts}       
\usepackage{nicefrac}       
\usepackage{microtype}      
\usepackage{xcolor}         

\usepackage{amsmath}    
\usepackage{amssymb}    

\usepackage{algorithm}   
\usepackage{algorithmic} 

\usepackage{wrapfig}
\usepackage{multirow} 
\usepackage[table]{xcolor}
\usepackage{graphicx}
\usepackage{cleveref}

\usepackage{capt-of}

\title{MemVLN: Episodic and Procedural Memory for Vision-and-Language Navigation}

\author{%
    Yuqi Liu$^{1}$ \hspace{1pt}
    Shengju Qian$^{2,\dagger}$ \hspace{1pt}
    Tianyuan Qu$^{1}$ \hspace{1pt}
    Mingxian Lin$^{3}$ \hspace{1pt}
    Zixuan Wang$^{1}$ \hspace{1pt} 
    \\
    \bfseries
    Xin Wang$^{2}$ \hspace{1pt}
    Bei Yu$^{1}$ \hspace{1pt}
    Jiaya Jia$^{4}$ \hspace{1pt}
    \\ 
    $^{1}$The Chinese University of Hong Kong \hspace{2pt} $^{2}$LIGHTSPEED \hspace{2pt}
    $^{3}$The University of Hong Kong \hspace{2pt} \\
    $^{4}$The Hong Kong University of Science and Technology \hspace{2pt} {\tt \small $^{\dagger}$Project Leader}   \\
}

\begin{document}

\maketitle

\begin{center}
    \includegraphics[width=1.\linewidth]{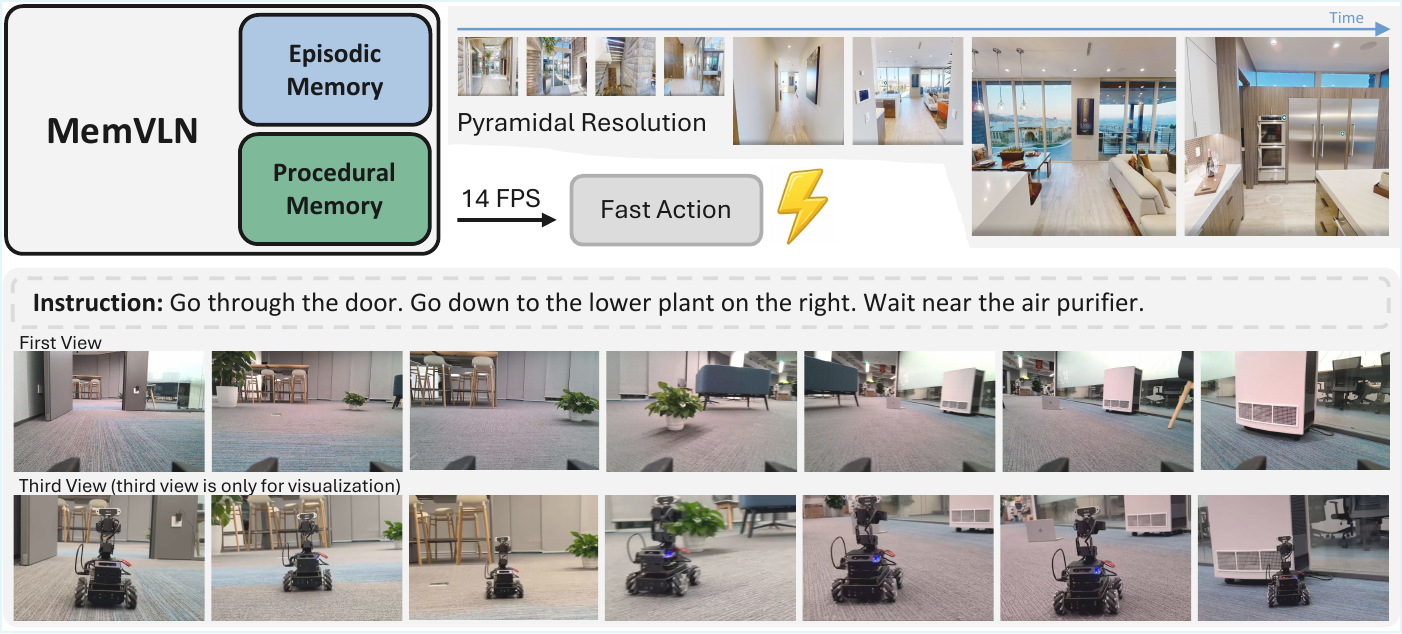}
    \captionof{figure}{MemVLN mimics human episodic memory and procedural memory. MemVLN employs pyramidal resolutions for the episodic memory to manage visual observations, and stores a set of mid-level actions in the procedural memory to enable fast inference.}
    \label{fig:teaser}
\end{center}

\begin{abstract}
  Vision-and-Language Navigation in Continuous Environments (VLN-CE) requires agents to maintain long-horizon visual history for trajectory consistency while executing actions with low latency. Existing video-based VLN approaches typically struggle to satisfy both demands simultaneously. To address these challenges, we propose MemVLN, a novel VLN framework that achieves state-of-the-art performance with real-time inference efficiency (14 FPS). MemVLN utilizes a visual encoder to process continuous observations and a Large Language Model (LLM) to interpret instructions and generate actions. Central to our approach is an Episodic Memory management that applies pyramidal resolutions. This mechanism concentrates computation on immediate percepts while retaining compressed long-term history. Complementing to this design, we introduce Procedural Memory for fast action with a compact vocabulary of atomic mid-level actions to bypass auto-regressive decoding latency. Experiments on VLN-CE show that MemVLN-4B surpasses the baseline Qwen3-VL-4B architecture by 5.8\% SR in R2R and 9.7\% SR in RxR, while achieving a 7$\times$ speedup in inference latency. 
\end{abstract}

\input{doc/1_intro}

\input{doc/2_related}

\input{doc/3_method}

\input{doc/4_experiment}

\input{doc/5_conclusion}

\clearpage
\bibliographystyle{plainnat}
\bibliography{references}


\appendix
\newpage

\input{doc/appendix}
\end{document}

%% file: doc/1_intro.tex
\section{Introduction}

Vision-and-language navigation (VLN) in continuous environments requires an agent to follow natural language instructions from a source to a target destination within unseen spaces, without the aid of prior maps. Unlike graph-based settings, this continuous setting (e.g., R2R-CE, RxR-CE) demands that agents ground linguistic cues directly into raw visual observations to plan low-level actions. While earlier approaches have made progress in narrowing the sim-to-real gap \citep{anderson2021sim,krantz2021waypoint,hong2022bridging}, these systems often struggle to generalize across domains due to their reliance on specific input modalities, such as depth sensors or precise odometry, which may be unreliable or unavailable. 

Recently, the emergence of large vision-and-language models (LVLMs) \citep{bai2025qwen3,li2024llava} has driven a shift toward unified vision-and-language navigation frameworks \citep{zhang2024navid,zhang2024uni,cheng2024navila,wei2025streamvln}. By integrating visual encoding, instruction understanding, and action prediction into a single end-to-end model, VLN models offer the potential for more robust, generalized navigation. However, deploying them in continuous embodied environments requires simultaneously maintaining long-horizon visual history for trajectory consistency and ensuring low-latency execution for real-time control. Satisfying these demands poses significant challenges for LVLMs. Current adaptations often fail to address both needs effectively. Some methods \citep{cheng2024navila} rely on fixed-size frames (e.g., uniformly sample only 8 frames), which compromises the temporal density necessary for consistent instruction following. Others \citep{zhang2024uni,wei2025streamvln} attempt to reduce computational costs by compressing visual tokens via pooling or merging, sacrificing spatial details necessary for precise navigation.

To address these challenges, we propose MemVLN, a novel VLN framework designed to reconcile the need for long-form video context with the requirement for real-time responsiveness. MemVLN utilizes a visual encoder to process continuous observations and a large language model (LLM) to interpret instructions and generate actions. 
The core of MemVLN is the bio-inspired memory mechanism that mimics human episodic and procedural memory to optimize both observation management and inference speed.
For episodic memory, we implement a pyramidal resolution strategy. This approach allocates high-resolution processing to immediate visual inputs for precise local navigation, while keeping a highly compressed summary of past observations to maintain global path consistency.
For procedural memory, we overcome the latency bottleneck caused by the auto-regressive nature of LLMs through a fast action mechanism. By mapping continuous control signals directly into a small vocabulary of atomic action primitives, this module bypasses slow auto-regressive decoding and enables rapid, real-time action prediction.
Consequently, MemVLN effectively leverages long-term visual history while maintaining a real-time 14 FPS inference speed. 

We validate our approach through comprehensive evaluations on two standard VLN-CE benchmarks: R2R-CE and RxR-CE. Despite relying exclusively on RGB video streams and natural language instructions, MemVLN establishes new state-of-the-art performance, significantly outperforming existing baselines. Furthermore, extensive ablation studies confirm the critical roles of both the hierarchical memory mechanism and the fast action tokenization in achieving robust, real-time control. To summarize, our main contributions are:

\begin{itemize}
\item We introduce MemVLN, a novel VLN model that effectively leverages long-form visual observations and boosts inference speed to 14 FPS. 
\item We propose the bio-inspired memory mechanism to efficiently process long-form visual history and reduce inference latency. Our design features a pyramidal resolution strategy to mimic episodic memory, alongside a fast action module to mimic procedural memory.
\item We validate our approach on the R2R and RxR benchmarks, where MemVLN achieves SoTA results relying exclusively on monocular RGB visual inputs.
\end{itemize}

%% file: doc/2_related.tex
\section{Related Works}

\subsection{Large Vision-and-Language Models}
Large vision-and-language models \citep{zhong2025lyra,bai2025qwen3,team2023gemini,achiam2023gpt} have recently emerged as a prominent research frontier, demonstrating remarkable efficacy across a diverse range of multi-modal tasks. Architecturally, these models typically comprise a vision encoder (e.g., ViT \citep{dosovitskiy2020image}) paired with a pre-trained large language model backbone (e.g., LLaMA \citep{touvron2023llama}). To bridge these distinct modalities, various feature fusion strategies have been proposed. For instance, Flamingo \citep{alayrac2022flamingo} introduces cross-attention layers to inject visual features into a frozen LLM. Alternatively, LLaVA \citep{liu2023visual} adopts a simpler yet highly effective paradigm, employing a linear projection layer to map visual tokens from a frozen CLIP \citep{radford2021learning} encoder directly into the word embedding space of LLaMA. Recent advancements, including open-source models like Qwen-VL \citep{bai2025qwen3} and proprietary systems such as GPT-4V \citep{achiam2023gpt} and Gemini \citep{team2023gemini}, have further scaled both training data and model parameters, achieving unprecedented capabilities in multi-modal reasoning and interleaved image-text understanding.

\subsection{Vision-and-Language Navigation}


Vision-Language Navigation (VLN) requires an agent to navigate based on textual instructions and visual cues. Initial efforts were constrained to discretized navigation graphs, where the agent's movement was limited to "teleporting" between predefined nodes~\citep{anderson2018vision,ku2020room,qi2020reverie,thomason2020vision,fried2018speaker,fu2020counterfactual,hong2020language}. Although these methods focus on high-level decision-making, they often overlook the intricacies of physical navigation. Consequently, VLN-CE ~\citep{krantz2020beyond,savva2019habitat} emerged as a more realistic benchmark, characterized by continuous action spaces and unconstrained movement. To address the challenges of continuous navigation, prior research generally follows two trajectories. One line of work directly predicts low-level control commands~\citep{chen2021topological,chen2022weakly,georgakis2022cross,raychaudhuri2021language,an2024etpnav,wang2023dreamwalker} to achieve fine-grained movement. Alternatively, hierarchical approaches utilize waypoint predictors~\citep{hong2022bridging,krantz2022sim,krantz2021waypoint,wang2023scaling,wang2024lookahead} to identify intermediate sub-goals, which are then reached via local controllers. However, a critical limitation of the latter is their heavy reliance on scene-specific features, which often leads to suboptimal generalization in novel, unseen environments.

\subsection{Large Models for Navigation}

Recent advancements in VLN have increasingly gravitated toward the integration of large models~\citep{zhong2025lyra,li2024llava,liu2023visual,bai2025qwen3}.
One prevalent paradigm~\citep{zhou2024navgpt,long2024discuss,long2024instructnav} leverages large language models (LLMs) as high-level heuristic planners within modular frameworks. However, these methods often exhibit a performance bottleneck when compared to specialized, task-specific architectures.
Conversely, an emerging line of research~\citep{cheng2024navila,zhang2024uni,zhang2024navid,wei2025streamvln} utilizes large vision-language models (LVLMs) ~\citep{li2024llama,lin2024vila} to capture complex spatio-temporal dynamics and generate low-level control commands in an end-to-end fashion. By harnessing cross-modal representations pre-trained on internet-scale data, these models inherit extensive world knowledge, thereby facilitating generalizable and scalable navigation in diverse environments. 
However, critical limitations remain in managing long visual contexts. While existing frameworks like Uni-Navid~\citep{zhang2024uni} and StreamVLN~\citep{wei2025streamvln} utilize token merging for sequence compression and memory control, these techniques destroy the uniform 2D spatial grid, creating fundamental conflicts with advanced techniques such as M-RoPE~\citep{wang2024qwen2}.

%% file: doc/3_method.tex





\begin{figure}[t]
    \centering
    \includegraphics[width=1.\linewidth]{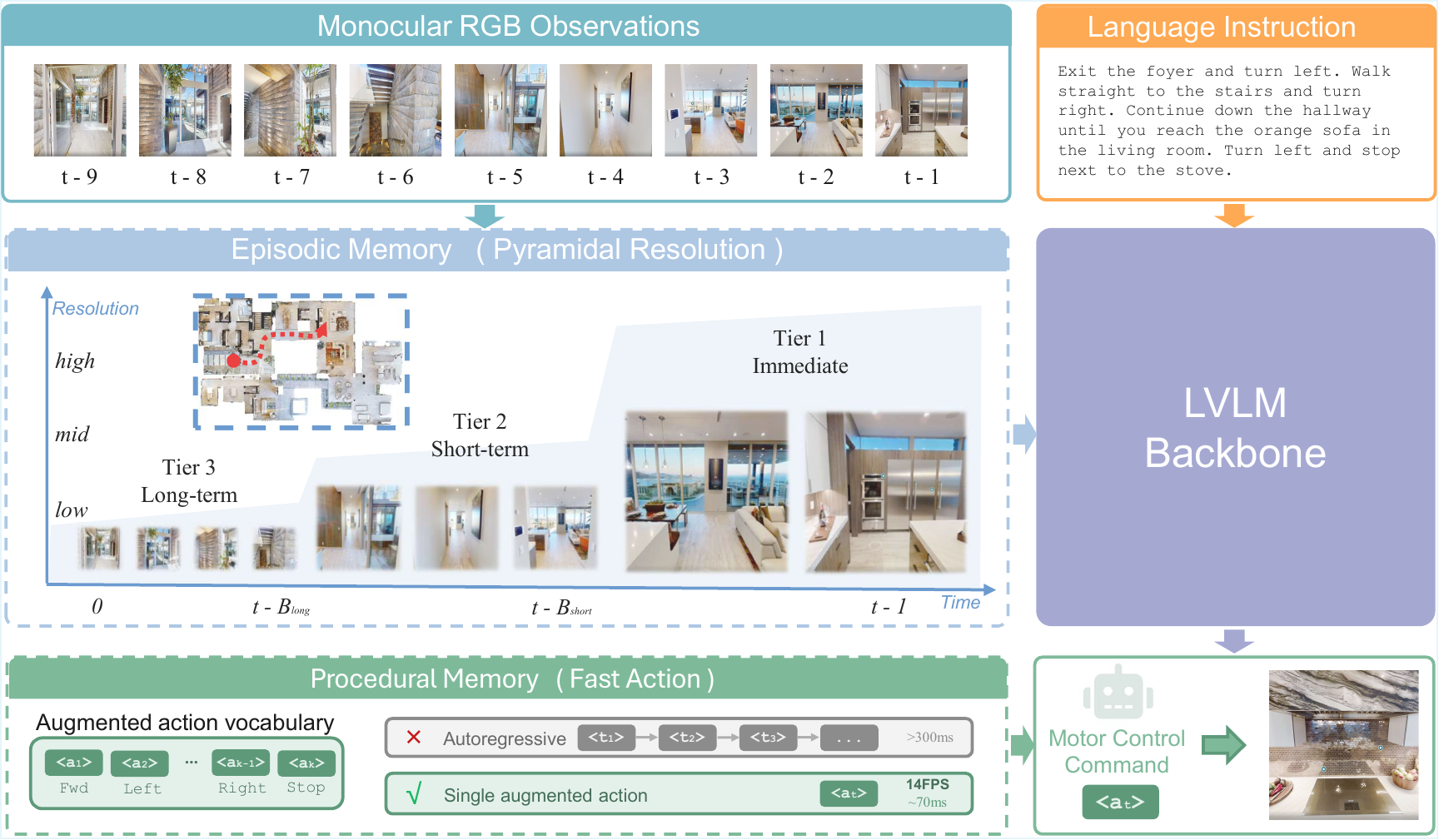}
    \caption{Overview of the MemVLN Architecture. Given a natural language instruction and monocular RGB observations, MemVLN utilizes an Episodic Memory module to construct pyramidal resolution observation. Complementarily, the Procedural Memory maintains a set of single-token mid-level actions to enable fast navigation.}
    \label{fig:architecture}
\end{figure}

\section{Method}
We present MemVLN, a VLN model for real-time, long-horizon navigation in VLN-CE. Following the architectural overview (Sec.~\ref{sec:arch}), we propose two novel memory designs: (i) the episodic memory for observations management (Sec.~\ref{sec:memory}), which utilizes pyramidal resolutions to process extensive visual histories; and (ii) the procedural memory for fast action (Sec.~\ref{sec:action}), a single-shot decoding strategy achieving 14 FPS. Together, these components empower MemVLN with both global trajectory awareness and fluid, reactive control capabilities. Finally, we detail the training objectives and the data used to optimize MemVLN (Sec.~\ref{sec:training}). 

\subsection{Task Definition and MemVLN}
\label{sec:arch}

\textbf{Task Definition.} 
Vision-and-Language Navigation in Continuous Environments (VLN-CE) benchmarks the capability of an embodied agent to execute cross-modal instruction following within unmapped 3D spaces. 
Given a natural language instruction $\mathcal{I}$, the agent must navigate from a source to a target location by processing an ongoing stream of egocentric RGB observations. 
At each discrete time step $t$, the agent receives a single instantaneous observation $v_{t-1} \in \mathbb{R}^{H \times W \times 3}$. 
By conditioning on the current percept $v_{t-1}$ and the historical context $\mathcal{H}_{t-1} = \{v_0, \dots, v_{t-2}\}$, the agent predicts an atomic action $a_t$ from a predefined set $\mathcal{A} = \{ \texttt{move\_forward}, \texttt{turn\_left}, \texttt{turn\_right}, \texttt{stop} \}$. 
Distinguishing itself from discrete graph-based navigation, the continuous setting 
bridges the gap between high-level semantic reasoning and low-level motor control.

 \textbf{MemVLN Architecture.} 
Given the complete sequence of historical frames and current frame $\mathcal{H}_t = \{v_0, \dots, v_{t-1}\}$, MemVLN first employs the pyramidal resolution mechanism to strategically select a subset $\mathcal{S}_t \subset \mathcal{H}_t$, effectively filtering out visual redundancies. 
These selected frames are then processed by a visual encoder $\mathcal{E}_{vis}$ to generate a sequence of visual features $\mathbf{F}_t = \mathcal{E}_{vis}(\mathcal{S}_t)$. 
To facilitate multi-modal decision-making, these visual features are integrated with the textual instruction $\mathcal{I}$ and fed into the Large Language Model (LLM) backbone. 
By leveraging the capabilities of the LLM, the model captures complex cross-modal dependencies between the visual path and linguistic landmarks. 
Finally, to bridge the gap between open-ended language generation and discrete navigation control, MemVLN defines a compacted vocabulary specifically tailored for fast navigation actions. 
The final action $a_t$ is then decoded from this optimized output space, ensuring efficient and precise policy execution. The whole architecture is illustrated in \Cref{fig:architecture}.

\subsection{Episodic Memory with Pyramidal Resolution}
\label{sec:memory}

A fundamental challenge in VLN-CE is the computational burden of modeling long-horizon visual observations, which are essential for maintaining path consistency and preventing trajectory drift. Inspired by episodic memory, a form of long-term conscious memory that enables individuals to recall past experiences within specific temporal and spatial contexts, we propose a pyramidal resolution module. This module emulates how biological systems selectively encode, store, and retrieve historical information, enabling efficient utilization of long-term visual observations without incurring prohibitive computational costs.

\begin{figure}[t]
    \centering
    \includegraphics[width=1.\linewidth]{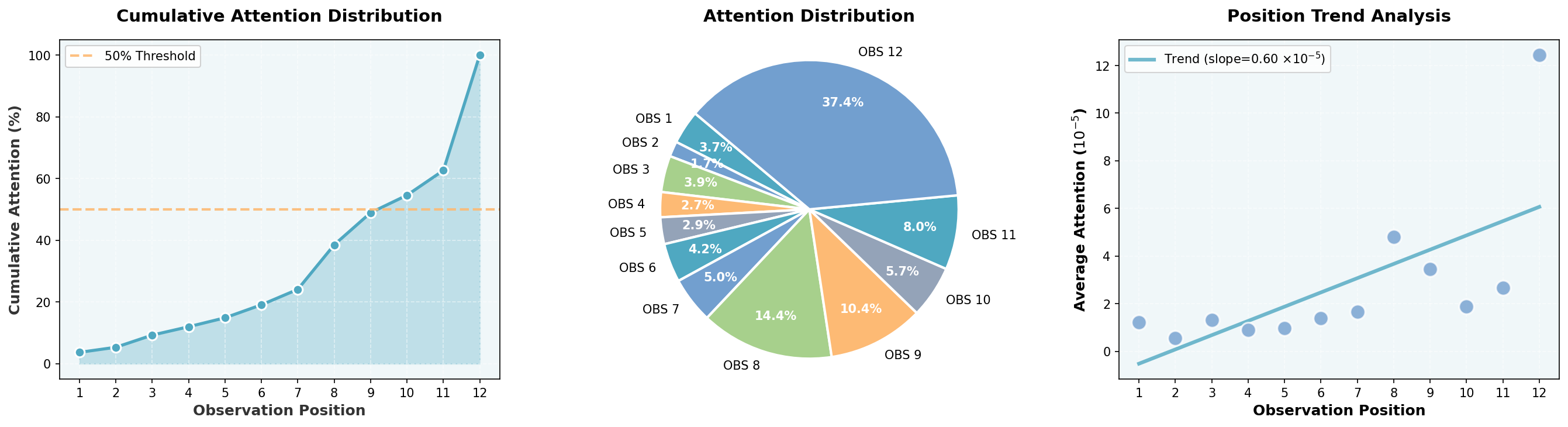}
    \caption{Attention analysis across visual observations. Left: The final four frames account for over 50\% of the total visual attention. Middle: The more recent observation receives higher concentration of attention. Right: Attention scores increase significantly as observations become more recent.}
    \label{fig:attn_analysis}
\end{figure}

 \textbf{Observation Attention Analysis.} 
Figure \ref{fig:attn_analysis} analyzes the distribution of observation attention weights during action generation. For this analysis, we randomly pick one step and uniformly sample 12 historical observations and calculate the average attention score for each observation. The leftmost plot shows that the final four observations account for more than 50\% of the total visual observation weight across all observations. In the middle pie chart, we observe that the most recent observation is weighted significantly higher than previous ones. Additionally, the scatter plot on the right illustrates that attention scores increase significantly for more recent observations. These findings suggest that in VLN tasks, recent observations play a more critical role and receive the majority of the model's focus, whereas earlier observations are largely disregarded.

\begin{wrapfigure}{r}{0.58\linewidth}  
    \vspace{-1em}
    \centering
    \includegraphics[width=\linewidth]{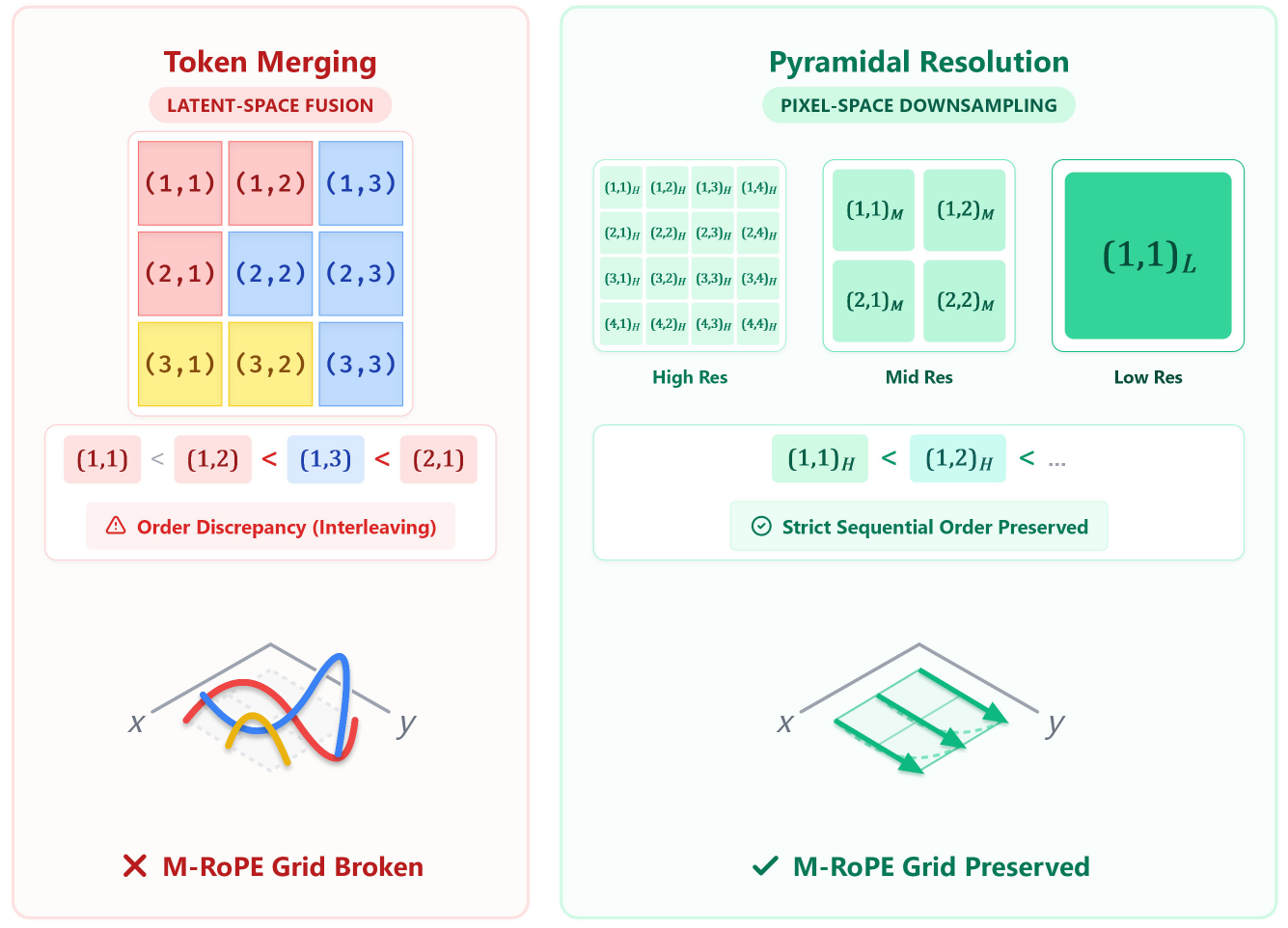}
    \caption{Pyramidal Resolution vs. Token Merging. Our pyramidal resolution ensures compatibility with modern LVLM design such as M-RoPE and DeepStack.}
    \label{fig:pyra_vs_tome}
    \vspace{-1em}
\end{wrapfigure}
\textbf{Preliminary: Why not Token Merging?}
While existing VLN frameworks like Uni-Navid \citep{zhang2024uni} and StreamVLN \citep{wei2025streamvln} utilize token merging for visual sequence compression and memory control, such approaches reveal substantial constraints within the context of modern architectural designs.
These latent-space methods cluster tokens based on semantic similarity rather than physical location. 
This irregular fusion destroys the uniform 2D spatial grid, causing fundamental conflicts with advanced positional encodings like M-RoPE \citep{wang2024qwen2} and disrupting the continuous sequence structures required by frameworks like DeepStack \citep{meng2024deepstack}. 
Conversely, our Pyramidal Resolution performs spatial downsampling in the raw pixel space before tokenization. By maintaining a strict, uniform spatial topology, our method seamlessly supports M-RoPE at any resolution and ensures full compatibility with DeepStack. It effectively compresses the context window without altering the internal mechanics of different deep transformer backbone.

\begin{figure}[t]
    \centering
    \includegraphics[width=1.\linewidth]{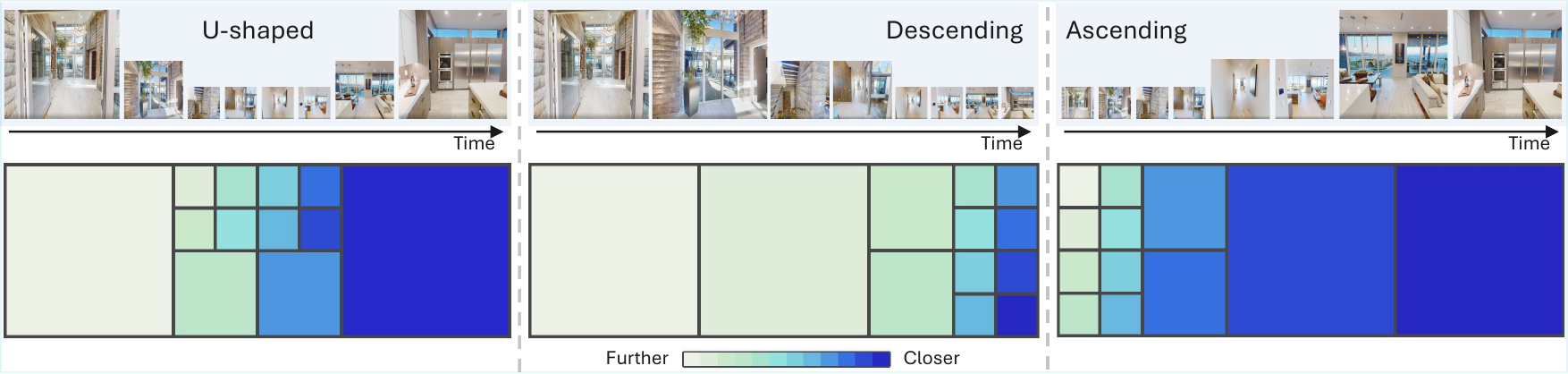}
    \caption{Top: Different types of pyramidal resolutions. Bottom: Attention distribution. Darker colors represent closer frames, which tend to receive more attention.}
    \label{fig:variants}
\end{figure}

\textbf{Pyramidal Resolution.} 
Human memory exhibits a temporal granularity bias, where recent events retain high perceptual fidelity while distant experiences undergo progressive spatial compression. Motivated by this cognitive mechanism and our preceding attention analysis, MemVLN introduces a pyramidal resolution strategy. 
Rather than treating all historical observations uniformly, we categorize frames into three temporal tiers: \textit{immediate}, \textit{short-term}, and \textit{long-term}, separated by two temporal boundaries, $B_{short}$ and $B_{long}$. 
For any frame $v_t$ in the history $\mathcal{H}_T$, its spatial resolution is adjusted via a rescaling function $\text{Rescale}(\cdot, r)$ based on its temporal distance from the current step $T$:

\begin{equation}
v'_t = 
\begin{cases} 
\text{Rescale}(v_t, r_{imm}), & T - B_{short} < t \leq T-1 \\
\text{Rescale}(v_t, r_{short}), & T - B_{long} < t \leq T - B_{short} \\
\text{Rescale}(v_t, r_{long}), & 0 \leq t \leq T - B_{long}
\end{cases}
\end{equation}

\noindent where $r_{imm} > r_{short} > r_{long}$ defines a descending hierarchy of resolutions. 
This pyramidal design is explicitly calibrated to maintain a balanced token distribution: the resolution $r_{long}$ and $r_{short}$ are set such that the cumulative tokens from the entire long-term sequence do not exceed the token count of a single immediate frame. 
By strategically downsampling distant observations, MemVLN significantly expands the agent's effective "look-back" window within the fixed context constraints, enabling the retention of sparse global landmarks without sacrificing the high-fidelity local cues necessary for immediate navigation.
We explore several variants for pyramidal resolution, as illustrated in \Cref{fig:variants}, including U-shaped, descending, and ascending. The descending prioritizes initial observations, whereas the ascending variant emphasizes the most recent ones. As a trade-off, the U-shaped design focuses on both the early and latest observations while down-sampling the intermediate steps.

\subsection{Procedural Memory for Fast Action}
\label{sec:action}

\begin{wraptable}{r}{0.28\textwidth}
\vspace{-2em}
\centering
\caption{Inference latency (ms) with different number of tokens.}
\vspace{1em}
\footnotesize
\label{tab:latency}
\setlength{\tabcolsep}{12pt}{
\begin{tabular}{@{}c|c@{}}
\toprule
\textbf{Tokens} & \textbf{Latency} \\ \midrule
1               & ~70                    \\
5               & 199                   \\
10              & 360                   \\ \bottomrule
\end{tabular}}
\vspace{-1em}
\end{wraptable}
\textbf{Latency Analysis.}
Real-time responsiveness is a critical prerequisite for embodied agents operating in continuous environments. While prior works have sought to enhance efficiency through mid-level actions \citep{cheng2024navila} or action chunking \citep{zhang2024uni}, our empirical analysis identifies the primary computational bottleneck as the inherent auto-regressive nature of VLMs. 
As shown in ~\Cref{tab:latency}, our latency computations reveal that generating the first token takes significantly longer due to pre-filling. Even with optimized KV-cache management, the subsequent token generation scales almost linearly with the output token count. Although a single-token forward pass is highly efficient, the iterative decoding required for complex commands or extended action sequences incurs delays that significantly exceed the time threshold necessary for real-time control.

\textbf{Fast Action.}
Autoregressive decoding introduces substantial latency in VLN, as actions are generated token by token through iterative inference. To overcome this bottleneck, we draw inspiration from procedural memory, a form of long-term implicit memory that enables the automatic execution of skills and motor behaviors. Based on this principle, we reformulate navigation as a one-shot mid-level action prediction task, allowing decisions to be produced in a single forward pass. 
By augmenting the output space with a specialized vocabulary of atomic mid-level action tokens, denoted as $\mathcal{A}_{aug}$, we enable each individual token to encapsulate either a mid-level action or a short-horizon action sequence. In practice, this is implemented by re-purposing a set of single tokens from the original LLM vocabulary to represent these augmented actions.
This design allows MemVLN to execute complex navigational decisions within a single forward pass, effectively bypassing the iterative decoding loop that typically bottlenecks real-time VLN systems. 
Consequently, our approach leverages the extreme inference efficiency of single-shot prediction to facilitate fluid, real-time deployment. The full augmented action token list $\mathcal{A}_{aug}$ is listed in the \textit{Appendix \Cref{sec:augmented_action}}.

\subsection{Training}
\label{sec:training}

\textbf{Objective.}
To optimize MemVLN, we use supervised fine-tuning that aligns multi-modal inputs with discrete navigational action. For each training sample at time step $t$, given the pyramidal visual features $\mathbf{F}_t$ and the linguistic instruction $\mathcal{I}$, the model is trained to maximize the likelihood of the ground-truth action $a_t^* \in \mathcal{A}_{aug}$, where $\mathcal{A}_{aug}$ denotes the augmented action list. 
Specifically, we employ a standard Cross-Entropy (CE) loss to supervise the action prediction. Since our fast action reduces the decision-making process to a single-shot prediction, the objective function $\mathcal{L}$ for a trajectory of length $T$ is formulated as:

\begin{equation}
    \mathcal{L} = - \frac{1}{T} \sum_{t=1}^{T} \log P(a_t^* \mid \mathbf{F}_t, \mathcal{I}; \theta) ,
\end{equation}

\noindent where $\theta$ represents the trainable parameters of MemVLN.

\textbf{Training Data.} 
We train MemVLN on a diverse corpus of navigation datasets. To construct this training set, we utilize R2R \citep{anderson2018vision}, RxR \citep{ku2020room}, and their augmented variants generated via EnvDrop \citep{tan2019learning}. Additionally, we incorporate randomly sampled 416K pairs of ScaleVLN \citep{wang2023scaling} sourced from the HM3D \citep{ramakrishnan2021habitat} and Gibson \citep{xia2018gibson} environments. 
To enhance the agent's adaptability to continuous environments, we further incorporate 557K human navigation trajectories \citep{cheng2024navila} and 278K synthetic data generated via the DAgger algorithm \citep{ross2011reduction}. 

%% file: doc/4_experiment.tex
\section{Experiment}
We evaluate MemVLN on the continuous VLN-CE benchmarks. First, we describe the experimental settings (Sec.~\ref{sec:setup}). Next, we compare MemVLN against other methods to demonstrate its superior performance on R2R and RxR (Sec.~\ref{sec:main_results}). We then conduct extensive ablation studies to verify the effectiveness of our pyramidal resolution and fast action (Sec.~\ref{sec:ablation}). 

\subsection{Experimental Settings}
\label{sec:setup}
\textbf{Simulation Benchmark Setup.} 
We evaluate our method on two standard VLN-CE benchmarks based on Matterport3D scenes within the Habitat simulator: R2R-CE [7] and RxR-CE [8]. While R2R-CE contains 5.6K English trajectories (avg. 10m), RxR-CE offers a more challenging scale with 126K multilingual instructions and longer, more complex trajectories (avg. 15m). 
Performance is assessed on the Val Unseen split. Following standard protocols, both utilize a 90° horizontal field-of-view (HFOV) and require agents to execute realistic navigation via continuous control. 

\textbf{Evaluation Metrics.} We evaluate navigation performance using navigation error (NE), oracle success rate (OS), success rate (SR), and success weighted by path length (SPL). 
Our evaluation protocol is consistent with prior work \citep{cheng2024navila}.

\textbf{Implementation Details.}
We adopt Qwen3-VL~\citep{bai2025qwen3} as the default backbone, and the maximum image resolution is set to $512 \times 512$. We employ a pyramidal resolution where immediate, short-term, and long-term frames are sampled at $512 \times 512$, $256 \times 256$, and $128 \times 128$, respectively.
The model is trained for a single epoch with a batch size of 128. We employ a cosine learning rate schedule with a peak learning rate of 1e-5 and a warmup ratio of 0.03. Training is conducted on a single server equipped with 8 GPUs, completing in approximately 27 hours. 

\begin{table}[t]
\centering
\caption{Comparison with state-of-the-art methods on VLN-CE R2R and RxR Val-Unseen split. $*$ indicates methods using the waypoint predictor from [34]. 'Pano.' stands for Panoramic Images, 'Odo.' stands for Odometry, and 'sRGB' refers to a single image.}
\footnotesize
\label{tab:results}
\setlength{\tabcolsep}{6pt} 
\begin{tabular}{l|cccc|cccc|ccc} 
\toprule
\multirow{2}{*}{\textbf{Method}} & \multicolumn{4}{c|}{\textbf{Observation}} & \multicolumn{4}{c|}{\textbf{R2R}} & \multicolumn{3}{c}{\textbf{RxR}} \\
\cmidrule(lr){2-5} \cmidrule(lr){6-9} \cmidrule(lr){10-12}  
 & Pano. & Odo. & Depth & sRGB & NE$\downarrow$ & OS$\uparrow$ & SR$\uparrow$ & SPL$\uparrow$ & NE$\downarrow$ & SR$\uparrow$ & SPL$\uparrow$ \\
\midrule
\multicolumn{12}{l}{\textbf{\textcolor{gray}{Waypoint-based Navigation}}} \\
CMA$^*$ \citep{hong2022bridging} & \checkmark & \checkmark & \checkmark & & 6.20 & 52.0 & 41.0 & 36.0 & 8.76 & 26.5 & 22.1 \\
VLN$\circlearrowright$BERT$^*$ \citep{hong2022bridging} & \checkmark & \checkmark & \checkmark & & 5.74 & 53.0 & 44.0 & 39.0 & 8.98 & 27.0 & 22.6 \\
Sim2Sim$^*$ \citep{krantz2022sim} & \checkmark & \checkmark & \checkmark & & 6.07 & 52.0 & 43.0 & 36.0 & - & - & - \\
GridMM$^*$ \citep{wang2023gridmm} & \checkmark & \checkmark & \checkmark & & 5.11 & 61.0 & 49.0 & 41.0 & - & - & - \\
ETPNav$^*$ \citep{an2024etpnav} & \checkmark & \checkmark & \checkmark & & \textbf{4.71} & 65.0 & 57.0 & 49.0 & 5.64 & 54.7 & 44.8 \\
ScaleVLN$^*$ \citep{wang2023scaling} & \checkmark & \checkmark & \checkmark & & 4.80 & - & 55.0 & 51.0 & - & - & - \\
\midrule
\multicolumn{12}{l}{\textbf{\textcolor{gray}{Continuous Action Control}}} \\
AG-CMTP \citep{chen2021topological} & \checkmark & \checkmark & \checkmark & & 7.90 & 39.2 & 23.1 & 19.1 & - & - & - \\
R2R-CMTP \citep{chen2021topological} & \checkmark & \checkmark & \checkmark & & 7.90 & 38.0 & 26.4 & 22.7 & - & - & - \\
LAW \citep{raychaudhuri2021language} & & \checkmark & \checkmark & \checkmark & 6.83 & 44.0 & 35.0 & 31.0 & 10.90 & 8.0 & 8.0 \\
CM2 \citep{georgakis2022cross} & & \checkmark & \checkmark & \checkmark & 7.02 & 41.5 & 34.3 & 27.6 & - & - & - \\
WS-MGMap \citep{chen2022weakly} & & \checkmark & \checkmark & \checkmark & 6.28 & 47.6 & 38.9 & 34.3 & - & - & - \\
ETPNav + FF \citep{wang2024sim} & & \checkmark & \checkmark & \checkmark & 5.95 & 55.8 & 44.9 & 30.4 & 8.79 & 25.5 & 18.1 \\
Seq2Seq \citep{krantz2020beyond} & & & \checkmark & \checkmark & 7.77 & 37.0 & 25.0 & 22.0 & 12.10 & 13.9 & 11.9 \\
CMA \citep{krantz2020beyond} & & & \checkmark & \checkmark & 7.37 & 40.0 & 32.0 & 30.0 & - & - & - \\
\midrule
\multicolumn{12}{l}{\textbf{\textcolor{gray}{Vision-and-Language Models (VLMs)}}} \\
NaVILA-8B \citep{cheng2024navila} & & & & \checkmark & 5.22 & 62.5 & 54.0 & 49.0 & 6.77 & 49.3 & 44.0 \\
Uni-NaVid-7B \citep{zhang2024uni} & & & & \checkmark & 5.58 & 53.3 & 47.0 & 42.7 & 6.24 & 48.7 & 40.9 \\
StreamVLN-7B \citep{wei2025streamvln} & & & & \checkmark & 4.98 & 64.2 & 56.9 & \textbf{51.9} & 6.22 & 52.9 & 46.0 \\
\rowcolor{gray!10} MemVLN-4B & & & & \checkmark & 5.34 & 63.2 & 56.6 & 50.0 & \textbf{4.22} & \textbf{66.5} & \textbf{57.4} \\
\rowcolor{gray!10} MemVLN-8B & & & & \checkmark & 4.98 & \textbf{65.3} & \textbf{58.4} & 51.2 & 4.56 & 66.0 & 57.3 \\
\bottomrule
\end{tabular}
\end{table}

\subsection{Main Results}
\label{sec:main_results}

We compare our approach against waypoint-based~\citep{krantz2021waypoint,hong2022bridging,krantz2022sim,wang2023gridmm,an2024etpnav,wang2023scaling}, continuous control~\citep{chen2021topological,raychaudhuri2021language,georgakis2022cross,chen2022weakly,wang2024sim,krantz2020beyond}, and VLM methods~\citep{cheng2024navila,zhang2024uni,wei2025streamvln}. These baselines utilize a variety of observation modalities, including panoramic images, odometry, and depth sensors. As shown in \Cref{tab:results},  MemVLN significantly outperforms methods that do not rely on simulator-pretrained waypoint predictors across both benchmarks using a single model.
Specifically, MemVLN-8B achieves state-of-the-art performance with 65.3\% OS and 58.4\% SR on R2R, and 66.0\% SR and 57.3\% SPL on RxR. These results demonstrate the robustness of our approach across both standard and long-horizon tasks. Notably, MemVLN is trained solely on single-view RGB input to outperform models utilizing richer observations, such as panoramic views or depth. This suggests that MemVLN effectively compensates for limited visual fields and sensor data through superior generalization. 
Interestingly, the model achieves better performance on RxR than on R2R. Qualitative analysis suggests that the detailed instructions in RxR provide stronger semantic grounding, keeping the agent on the correct path. 
Further qualitative details are provided in \Cref{sec:qualitative}. 
Following NaVILA \cite{cheng2024navila}, additional experiments trained exclusively on R2R yield a zero-shot performance of 47.9 OS on RxR, demonstrating cross-dataset generalization.

\subsection{Ablation Studies}
\label{sec:ablation}

We conduct a series of ablation studies to evaluate the effectiveness of our design and the impact of key hyperparameters. Unless otherwise specified, all models are initialized with Qwen3-VL-4B and trained exclusively on the standard training splits of R2R and RxR. Furthermore, we investigate the influence of data scale by comparing performance across different training data configurations.

\begin{table}[t]
\centering
\caption{Performance and inference speed comparison with and without fast action.}
\footnotesize
\label{tab:fast_action}
\setlength{\tabcolsep}{8pt} 
\begin{tabular}{l|c|cccc|cccc|c} 
\toprule
\multirow{2}{*}{\textbf{Method}} & \multirow{2}{*}{\textbf{FastAct}} & \multicolumn{4}{c|}{\textbf{R2R}} & \multicolumn{4}{c|}{\textbf{RxR}} & \multirow{2}{*}{\textbf{FPS$\uparrow$}} \\
\cmidrule(lr){3-6} \cmidrule(lr){7-10} 
 & & NE$\downarrow$ & OS$\uparrow$ & SR$\uparrow$ & SPL$\uparrow$ & NE$\downarrow$ & OS$\uparrow$ & SR$\uparrow$ & SPL$\uparrow$ & \\
\midrule
MemVLN-4B        &            & 5.7 & 58.9 & 50.8 & 45.6 & 5.3 & 65.0 & 56.8 & 49.3 & ~2 \\
MemVLN-4B        & \checkmark & \textbf{5.3} & \textbf{59.9} & \textbf{52.6} & \textbf{47.7} & \textbf{5.2} & \textbf{66.7} & \textbf{58.9} & \textbf{51.7} & 14 \\
\bottomrule
\end{tabular}
\end{table}

\textbf{Procedural Memory for Fast Action.}
In Table \ref{tab:fast_action}, we ablate the performance and inference efficiency of our procedural fast action. We compare our approach against the textual mid-level actions of NaVILA~\citep{cheng2024navila}, which predicts descriptive text sequences (e.g., `The next action is turn left 30 degrees'). 
By contrast, MemVLN circumvents the auto-regressive decoding bottleneck by mapping atomic mid-level actions to specialized single tokens. 
We evaluate the proposed fast action against standard auto-regressive decoding under an identical 8-frame uniform sampling protocol. The empirical results demonstrate two significant advantages. The most immediate benefit is a drastic reduction in inference latency. By mapping mid-level actions to specialized single tokens, the model bypasses sequential decoding bottlenecks, achieving a 7$\times$ speedup (accelerating from 2 FPS to a real-time 14 FPS). 
Besides, it consistently improves navigation accuracy across multiple benchmarks. Specifically, the model yields absolute gains of 1.0 OS and 1.8 SR on R2R, and 1.7 OS and 2.1 SR on RxR.

\begin{table}[htbp]
\centering

\caption{Performance comparison across different pyramidal configurations. The BEST column denotes the count of best-performing metrics per row.}
\label{tab:pyramidal_config}
\footnotesize
\setlength{\tabcolsep}{8pt} 
\begin{tabular}{ccc|cccc|cccc|c} 
\toprule
\multicolumn{3}{c|}{\textbf{Pyramidal Config}} & \multicolumn{4}{c|}{\textbf{R2R}} & \multicolumn{4}{c|}{\textbf{RxR}} & \textbf{BEST} \\
\cmidrule(r){1-3} \cmidrule(lr){4-7} \cmidrule(lr){8-11} \cmidrule(l){12-12}
\# imm & \# short & \# long & NE$\downarrow$ & OS$\uparrow$ & SR$\uparrow$ & SPL$\uparrow$ & NE$\downarrow$ & OS$\uparrow$ & SR$\uparrow$ & SPL$\uparrow$ & CNT \\
\midrule
8 & 0 & 0  & \textbf{5.3} & \textbf{59.9} & 52.6 & 47.7 & 5.2 & \textbf{66.7} & 58.9 & 51.7 & 3 \\
6 & 4 & 16 & \textbf{5.3} & 59.2 & \textbf{54.2} & \textbf{49.6} & 5.1 & 66.4 & 60.3 & \textbf{53.0} & 4 \\
4 & 8 & 32 & \textbf{5.3} & 59.7 & 53.8 & \textbf{49.6} & \textbf{4.9} & 66.2 & \textbf{60.4} & \textbf{53.0} & \textbf{5} \\
\bottomrule
\end{tabular}

\vspace{1.5em} 

\caption{Performance comparison across different pyramidal types.}
\label{tab:pyramidal_types}
\footnotesize
\setlength{\tabcolsep}{12pt} 
\begin{tabular}{l|cccc|cccc} 
\toprule
\multirow{2}{*}{\textbf{Pyramidal Types}} & \multicolumn{4}{c|}{\textbf{R2R}} & \multicolumn{4}{c}{\textbf{RxR}} \\
\cmidrule(lr){2-5} \cmidrule(lr){6-9}
 & NE$\downarrow$ & OS$\uparrow$ & SR$\uparrow$ & SPL$\uparrow$ & NE$\downarrow$ & OS$\uparrow$ & SR$\uparrow$ & SPL$\uparrow$ \\
\midrule
U-Shaped & 5.6 & 57.1 & 50.9 & 47.3 & 5.1 & 65.6 & 60.0 & 52.5 \\
Descending & 6.3 & 51.3 & 43.8 & 39.7 & 6.7 & 54.0 & 46.8 & 40.5 \\
Ascending & \textbf{5.3} & \textbf{59.7} & \textbf{53.8} & \textbf{49.6} & \textbf{4.9} & \textbf{66.2} & \textbf{60.4} & \textbf{53.0} \\
\bottomrule
\end{tabular}

\vspace{1.5em} 

\caption{Ablation study on training data.}
\footnotesize
\label{tab:training_data}
\setlength{\tabcolsep}{5pt}
\begin{tabular}{ccccc|cccc|cccc}
\toprule
\multirow{2}{*}{\textbf{VLN-CE}} & \multirow{2}{*}{\textbf{Human}} & \multirow{2}{*}{\textbf{Scale}} & \multirow{2}{*}{\textbf{DAgger}} & \multirow{2}{*}{\textbf{Envdrop}} & \multicolumn{4}{c|}{\textbf{R2R Val-Unseen}} & \multicolumn{4}{c}{\textbf{RxR Val-Unseen}} \\
\cmidrule(lr){6-9} \cmidrule(lr){10-13}
& & & & & NE$\downarrow$ & OS$\uparrow$ & SR$\uparrow$ & SPL$\uparrow$ & NE$\downarrow$ & OS$\uparrow$ & SR$\uparrow$ & SPL$\uparrow$ \\
\midrule
$\checkmark$ & & & & & 5.3 & 59.7 & 53.8 & 49.6 & 4.9 & 66.2 & 60.4 & 53.0 \\
$\checkmark$ & $\checkmark$ & & & & 5.4 & 58.6 & 52.6 & 48.9 & 5.4 & 63.4 & 57.4 & 51.2 \\
$\checkmark$ & $\checkmark$ & $\checkmark$ & & & 5.2 & 59.7 & 54.6 & 50.6 & 5.2 & 63.3 & 58.2 & 51.4 \\
$\checkmark$ & $\checkmark$ & $\checkmark$ & $\checkmark$ & & 5.4 & 61.0 & 54.6 & 48.1 & 4.7 & 70.8 & 63.9 & 54.2 \\
$\checkmark$ & $\checkmark$ & $\checkmark$ & $\checkmark$ & $\checkmark$ & 5.3 & 63.2 & 56.6 & 50.0 & 4.2 & 72.8 & 66.5 & 57.4 \\
\bottomrule
\end{tabular}

\end{table}

\textbf{Episodic Memory for Observations Management.}
We employ a pyramidal resolution where immediate, short-term, and long-term frames are sampled at $512 \times 512$, $256 \times 256$, and $128 \times 128$, respectively. This multi-scale design allows the model to capture an extended visual history without increasing the total visual token budget. To validate this approach, we conduct ablation studies across various pyramidal configurations while maintaining a constant number of visual tokens. 
As shown in \Cref{tab:pyramidal_config}, the model equipped with pyramidal resolution outperforms the baseline without it. Specifically, the most notable improvements are observed in SR and SPL, demonstrating absolute gains of 1.2 in SR and 1.9 in SPL on the R2R, as well as 1.5 in SR and 1.3 in SPL on RxR. This indicates that the pyramidal design effectively enhances both navigation accuracy and path efficiency, with the (4, 8, 32) configuration achieving the best overall performance. Furthermore, we find that the OS rate remains largely unchanged. This suggests that while models without the pyramidal design can successfully pass through the target location, they fail to execute the correct stopping action.
As shown in Table~\ref{tab:pyramidal_types}, the ascending pyramidal configuration achieves the best performance. This aligns with our expectations, as recent observations typically carry the most critical information for immediate decision-making, making a higher resolution for these recent inputs highly beneficial.

\textbf{Training Data.} 
We conduct an ablation study on training data configurations in Table \ref{tab:training_data}. 
Our results show that 
performance scales with the training corpus size. 
Interestingly, incorporating human activity and ScaleVLN data leads to a marginal performance drop; however, we retain these subsets to enhance environmental diversity, as R2R and RxR are restricted to indoor trajectories. 
Notably, DAgger data yields significant gains, improving OS by 2.4\% and SR by 0.6\% on R2R, and 4.6\% OS / 3.5\% SR on RxR over the baseline, 
which underscores the importance of interactive, on-policy data for robust navigation. Finally, the inclusion of all training data provides the best performance.

%% file: doc/5_conclusion.tex
\section{Conclusion}
This paper introduces MemVLN, a novel VLN framework that achieves state-of-the-art performance in VLN-CE tasks with real-time inference. By introducing pyramidal resolutions, we effectively balance the retention of long-horizon visual history with computational efficiency. Furthermore, our fast action significantly reduces the latency typically associated with auto-regressive decoding.
Extensive evaluations on R2R and RxR benchmarks demonstrate that MemVLN outperforms existing methods in terms of successful rate and latency, demonstrating a scalable and robust path for deploying VLN models in continuous environments. 

%% file: doc/appendix.tex
\section{Limitations}
The limitations of this work stem from the computational and hardware requirements. First, scaling this architecture to massive, trillion-parameter models is currently constrained by computational resources. Second, while MemVLN achieves real-time inference, it still relies on GPU acceleration, which may limit deployment on resource-constrained platforms, such as lightweight edge devices or mobile robots with strictly limited power budgets. Future work will explore model quantization and edge-optimized deployment to address these hardware constraints.

\section{Compute Resources}
All experiments were conducted on a server equipped with eight NVIDIA H200 GPUs, each with 140 GB of memory. Training on the full dataset took approximately 27 hours. 

\section{Cross-dataset Generalization}
To evaluate cross-dataset generalization, we follow NaVILA \cite{cheng2024navila} and train MemVLN exclusively on the R2R dataset. We then evaluate its zero-shot performance on the RxR Val-Unseen split. As Table \ref{tab:rxr_zeroshot} shows, our method outperforms NaVILA, the current state-of-the-art, improving SR by 12\% and SPL by 20\%.

\begin{table}[t]
\centering
\caption{Cross-dataset performance on the RxR-CE Val Unseen split. MemVLN is trained exclusively on R2R training split in this setting.}
\footnotesize
\label{tab:rxr_zeroshot}
\setlength{\tabcolsep}{9pt} 
\begin{tabular}{l|ccc|cccc}
\toprule
\multirow{2}{*}{\textbf{Method}} & \multicolumn{3}{c}{\textbf{Observation}} & \multicolumn{4}{c}{\textbf{RxR Val-Unseen}} \\
\cmidrule(lr){2-4} \cmidrule(lr){5-8}
 & S.RGB & Depth & Odo. & NE $\downarrow$ & OS $\uparrow$ & SR $\uparrow$ & SPL $\uparrow$ \\
\midrule
LAW \cite{raychaudhuri2021language} & \checkmark & \checkmark & \checkmark & 10.87 & 21.0 & 8.0 & 8.0 \\
CM2 \cite{georgakis2022cross} & \checkmark & \checkmark & \checkmark & 8.98 & 25.3 & 14.4 & 9.2 \\
WS-MGMap \cite{chen2022weakly} & \checkmark & \checkmark & \checkmark & 9.83 & 29.8 & 15.0 & 12.1 \\
Seq2Seq \cite{krantz2020beyond} & \checkmark & \checkmark & & 11.8 & 5.02 & 3.51 & 3.43 \\
CMA \cite{krantz2020beyond} & \checkmark & \checkmark & & 11.7 & 10.7 & 4.41 & 2.47 \\
RGB-Seq2Seq \cite{zhang2024navid} & \checkmark & & & 11.2 & 12.2 & 0.0 & 0.0 \\
RGB-CMA \cite{zhang2024navid} & \checkmark & & & 9.55 & 14.8 & 0.0 & 0.0 \\
\midrule
NaVid-7B \cite{zhang2024navid} & \checkmark & & & 8.41 & 34.5 & 23.8 & 21.2 \\
NaVILA-8B \cite{cheng2024navila} & \checkmark & & & 8.78 & 46.8 & 34.3 & 28.2 \\
\rowcolor{gray!10} MemVLN-4B & \checkmark & & & \textbf{7.71} & \textbf{47.9} & \textbf{38.4} & \textbf{33.8} \\
\bottomrule
\end{tabular}
\end{table}

\begin{figure}[t]
    \centering
    \includegraphics[width=1.\linewidth]{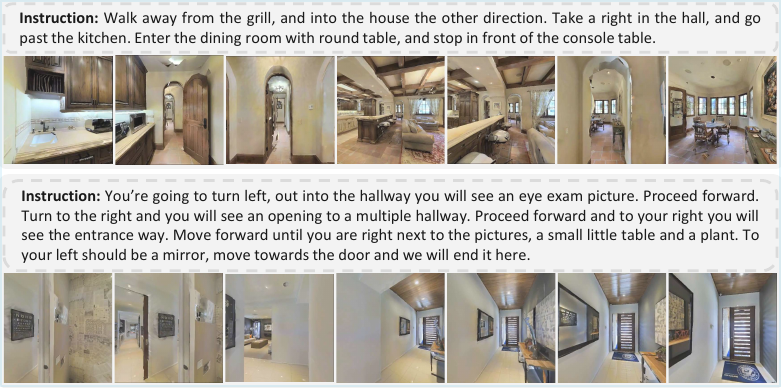}
    \caption{Visualization on R2R-CE and RxR-CE in the Habitat simulator.}
    \label{fig:visual_sim}
\end{figure}

\section{Qualitative Results}
\label{sec:qualitative}
\Cref{fig:visual_sim} illustrates qualitative navigation results within the Habitat simulator~\cite{savva2019habitat}. These visualizations confirm that MemVLN maintains high trajectory fidelity and temporal path consistency across diverse test episodes. 




\section{Augmented Action Token Vocabulary}
\label{sec:augmented_action}
By default, there are only four actions in VLN tasks: $\mathcal{A} = \{ \texttt{move\_forward},$ $ \texttt{turn\_left}, \texttt{turn\_right}, \texttt{stop} \}$. Predicting actions step-by-step is inefficient. Previous works have tried to speed up this process by predicting mid-level actions \cite{cheng2024navila} or using action chunking \cite{zhang2024uni}. However, these methods are still restricted by the auto-regressive nature of VLMs, as discussed in Section 3.3. Instead, our MemVLN augments the output space with a specialized vocabulary of atomic mid-level action tokens, denoted as $\mathcal{A}_{aug}$, which are listed in \Cref{tab:augmented_actions}. 

\begin{table}[t]
    \setlength{\tabcolsep}{15pt}
    \centering
    \caption{Augmented atomic mid-level action tokens ($\mathcal{A}_{aug}$) used in MemVLN.}
    \label{tab:augmented_actions}
    \begin{tabular}{c|l}
        \toprule
        \textbf{ID} & \textbf{Action Description} \\
        \midrule
        0 & Stop / End navigation \\
        1 & Move forward short distance (25cm) \\
        2 & Move forward medium distance (50cm) \\
        3 & Move forward long distance (75cm) \\
        4 & Turn left $15^\circ$ \\
        5 & Turn left $30^\circ$ \\
        6 & Turn left $45^\circ$ \\
        7 & Turn right $15^\circ$ \\
        8 & Turn right $30^\circ$ \\
        9 & Turn right $45^\circ$ \\
        \bottomrule
    \end{tabular}
\end{table}

%% file: references.bib
@article{meng2024deepstack,
  title={Deepstack: Deeply stacking visual tokens is surprisingly simple and effective for lmms},
  author={Meng, Lingchen and Yang, Jianwei and Tian, Rui and Dai, Xiyang and Wu, Zuxuan and Gao, Jianfeng and Jiang, Yu-Gang},
  journal={Advances in Neural Information Processing Systems},
  volume={37},
  pages={23464--23487},
  year={2024}
}

@article{wang2024qwen2,
  title={Qwen2-vl: Enhancing vision-language model's perception of the world at any resolution},
  author={Wang, Peng and Bai, Shuai and Tan, Sinan and Wang, Shijie and Fan, Zhihao and Bai, Jinze and Chen, Keqin and Liu, Xuejing and Wang, Jialin and Ge, Wenbin and others},
  journal={arXiv preprint arXiv:2409.12191},
  year={2024}
}

@inproceedings{xia2018gibson,
  title={Gibson env: Real-world perception for embodied agents},
  author={Xia, Fei and Zamir, Amir R and He, Zhiyang and Sax, Alexander and Malik, Jitendra and Savarese, Silvio},
  booktitle={Proceedings of the IEEE conference on computer vision and pattern recognition},
  pages={9068--9079},
  year={2018}
}

@article{ramakrishnan2021habitat,
  title={Habitat-matterport 3d dataset (hm3d): 1000 large-scale 3d environments for embodied ai},
  author={Ramakrishnan, Santhosh K and Gokaslan, Aaron and Wijmans, Erik and Maksymets, Oleksandr and Clegg, Alex and Turner, John and Undersander, Eric and Galuba, Wojciech and Westbury, Andrew and Chang, Angel X and others},
  journal={arXiv preprint arXiv:2109.08238},
  year={2021}
}

@article{team2023gemini,
  title={Gemini: a family of highly capable multimodal models},
  author={Team, Gemini and Anil, Rohan and Borgeaud, Sebastian and Alayrac, Jean-Baptiste and Yu, Jiahui and Soricut, Radu and Schalkwyk, Johan and Dai, Andrew M and Hauth, Anja and Millican, Katie and others},
  journal={arXiv preprint arXiv:2312.11805},
  year={2023}
}

@article{achiam2023gpt,
  title={Gpt-4 technical report},
  author={Achiam, Josh and Adler, Steven and Agarwal, Sandhini and Ahmad, Lama and Akkaya, Ilge and Aleman, Florencia Leoni and Almeida, Diogo and Altenschmidt, Janko and Altman, Sam and Anadkat, Shyamal and others},
  journal={arXiv preprint arXiv:2303.08774},
  year={2023}
}

@inproceedings{radford2021learning,
  title={Learning transferable visual models from natural language supervision},
  author={Radford, Alec and Kim, Jong Wook and Hallacy, Chris and Ramesh, Aditya and Goh, Gabriel and Agarwal, Sandhini and Sastry, Girish and Askell, Amanda and Mishkin, Pamela and Clark, Jack and others},
  booktitle={International conference on machine learning},
  pages={8748--8763},
  year={2021},
  organization={PmLR}
}

@article{alayrac2022flamingo,
  title={Flamingo: a visual language model for few-shot learning},
  author={Alayrac, Jean-Baptiste and Donahue, Jeff and Luc, Pauline and Miech, Antoine and Barr, Iain and Hasson, Yana and Lenc, Karel and Mensch, Arthur and Millican, Katherine and Reynolds, Malcolm and others},
  journal={Advances in neural information processing systems},
  volume={35},
  pages={23716--23736},
  year={2022}
}

@article{touvron2023llama,
  title={Llama: Open and efficient foundation language models},
  author={Touvron, Hugo and Lavril, Thibaut and Izacard, Gautier and Martinet, Xavier and Lachaux, Marie-Anne and Lacroix, Timoth{\'e}e and Rozi{\`e}re, Baptiste and Goyal, Naman and Hambro, Eric and Azhar, Faisal and others},
  journal={arXiv preprint arXiv:2302.13971},
  year={2023}
}

@article{dosovitskiy2020image,
  title={An image is worth 16x16 words: Transformers for image recognition at scale},
  author={Dosovitskiy, Alexey and Beyer, Lucas and Kolesnikov, Alexander and Weissenborn, Dirk and Zhai, Xiaohua and Unterthiner, Thomas and Dehghani, Mostafa and Minderer, Matthias and Heigold, Georg and Gelly, Sylvain and others},
  journal={arXiv preprint arXiv:2010.11929},
  year={2020}
}

@inproceedings{anderson2021sim,
  title={Sim-to-real transfer for vision-and-language navigation},
  author={Anderson, Peter and Shrivastava, Ayush and Truong, Joanne and Majumdar, Arjun and Parikh, Devi and Batra, Dhruv and Lee, Stefan},
  booktitle={Conference on Robot Learning},
  pages={671--681},
  year={2021},
  organization={PMLR}
}

@article{wang2024sim,
  title={Sim-to-real transfer via 3d feature fields for vision-and-language navigation},
  author={Wang, Zihan and Li, Xiangyang and Yang, Jiahao and Liu, Yeqi and Jiang, Shuqiang},
  journal={arXiv preprint arXiv:2406.09798},
  year={2024}
}

@inproceedings{wang2023gridmm,
  title={Gridmm: Grid memory map for vision-and-language navigation},
  author={Wang, Zihan and Li, Xiangyang and Yang, Jiahao and Liu, Yeqi and Jiang, Shuqiang},
  booktitle={Proceedings of the IEEE/CVF International conference on computer vision},
  pages={15625--15636},
  year={2023}
}

@inproceedings{ross2011reduction,
  title={A reduction of imitation learning and structured prediction to no-regret online learning},
  author={Ross, St{\'e}phane and Gordon, Geoffrey and Bagnell, Drew},
  booktitle={Proceedings of the fourteenth international conference on artificial intelligence and statistics},
  pages={627--635},
  year={2011},
  organization={JMLR Workshop and Conference Proceedings}
}

@inproceedings{tan2019learning,
  title={Learning to navigate unseen environments: Back translation with environmental dropout},
  author={Tan, Hao and Yu, Licheng and Bansal, Mohit},
  booktitle={Proceedings of the 2019 Conference of the North American Chapter of the Association for Computational Linguistics: Human Language Technologies, Volume 1 (Long and Short Papers)},
  pages={2610--2621},
  year={2019}
}

@article{wei2025streamvln,
  title={Streamvln: Streaming vision-and-language navigation via slowfast context modeling},
  author={Wei, Meng and Wan, Chenyang and Yu, Xiqian and Wang, Tai and Yang, Yuqiang and Mao, Xiaohan and Zhu, Chenming and Cai, Wenzhe and Wang, Hanqing and Chen, Yilun and others},
  journal={arXiv preprint arXiv:2507.05240},
  year={2025}
}

@article{zhang2024navid,
  title={Navid: Video-based vlm plans the next step for vision-and-language navigation},
  author={Zhang, Jiazhao and Wang, Kunyu and Xu, Rongtao and Zhou, Gengze and Hong, Yicong and Fang, Xiaomeng and Wu, Qi and Zhang, Zhizheng and Wang, He},
  journal={arXiv preprint arXiv:2402.15852},
  year={2024}
}

@inproceedings{lin2024vila,
  title={Vila: On pre-training for visual language models},
  author={Lin, Ji and Yin, Hongxu and Ping, Wei and Molchanov, Pavlo and Shoeybi, Mohammad and Han, Song},
  booktitle={Proceedings of the IEEE/CVF conference on computer vision and pattern recognition},
  pages={26689--26699},
  year={2024}
}

@inproceedings{li2024llama,
  title={Llama-vid: An image is worth 2 tokens in large language models},
  author={Li, Yanwei and Wang, Chengyao and Jia, Jiaya},
  booktitle={European Conference on Computer Vision},
  pages={323--340},
  year={2024},
  organization={Springer}
}

@article{long2024instructnav,
  title={Instructnav: Zero-shot system for generic instruction navigation in unexplored environment},
  author={Long, Yuxing and Cai, Wenzhe and Wang, Hongcheng and Zhan, Guanqi and Dong, Hao},
  journal={arXiv preprint arXiv:2406.04882},
  year={2024}
}

@inproceedings{long2024discuss,
  title={Discuss before moving: Visual language navigation via multi-expert discussions},
  author={Long, Yuxing and Li, Xiaoqi and Cai, Wenzhe and Dong, Hao},
  booktitle={2024 IEEE International Conference on Robotics and Automation (ICRA)},
  pages={17380--17387},
  year={2024},
  organization={IEEE}
}

@inproceedings{zhou2024navgpt,
  title={Navgpt: Explicit reasoning in vision-and-language navigation with large language models},
  author={Zhou, Gengze and Hong, Yicong and Wu, Qi},
  booktitle={Proceedings of the AAAI Conference on Artificial Intelligence},
  volume={38},
  number={7},
  pages={7641--7649},
  year={2024}
}

@article{bai2025qwen3,
  title={Qwen3-vl technical report},
  author={Bai, Shuai and Cai, Yuxuan and Chen, Ruizhe and Chen, Keqin and Chen, Xionghui and Cheng, Zesen and Deng, Lianghao and Ding, Wei and Gao, Chang and Ge, Chunjiang and others},
  journal={arXiv preprint arXiv:2511.21631},
  year={2025}
}

@article{liu2023visual,
  title={Visual instruction tuning},
  author={Liu, Haotian and Li, Chunyuan and Wu, Qingyang and Lee, Yong Jae},
  journal={Advances in neural information processing systems},
  volume={36},
  pages={34892--34916},
  year={2023}
}

@article{li2024llava,
  title={Llava-onevision: Easy visual task transfer},
  author={Li, Bo and Zhang, Yuanhan and Guo, Dong and Zhang, Renrui and Li, Feng and Zhang, Hao and Zhang, Kaichen and Zhang, Peiyuan and Li, Yanwei and Liu, Ziwei and others},
  journal={arXiv preprint arXiv:2408.03326},
  year={2024}
}

@inproceedings{zhong2025lyra,
  title={Lyra: An efficient and speech-centric framework for omni-cognition},
  author={Zhong, Zhisheng and Wang, Chengyao and Liu, Yuqi and Yang, Senqiao and Tang, Longxiang and Zhang, Yuechen and Li, Jingyao and Qu, Tianyuan and Li, Yanwei and Chen, Yukang and others},
  booktitle={Proceedings of the IEEE/CVF International Conference on Computer Vision},
  pages={3694--3704},
  year={2025}
}

@inproceedings{wang2024lookahead,
  title={Lookahead exploration with neural radiance representation for continuous vision-language navigation},
  author={Wang, Zihan and Li, Xiangyang and Yang, Jiahao and Liu, Yeqi and Hu, Junjie and Jiang, Ming and Jiang, Shuqiang},
  booktitle={Proceedings of the IEEE/CVF conference on computer vision and pattern recognition},
  pages={13753--13762},
  year={2024}
}

@inproceedings{wang2023scaling,
  title={Scaling data generation in vision-and-language navigation},
  author={Wang, Zun and Li, Jialu and Hong, Yicong and Wang, Yi and Wu, Qi and Bansal, Mohit and Gould, Stephen and Tan, Hao and Qiao, Yu},
  booktitle={Proceedings of the IEEE/CVF international conference on computer vision},
  pages={12009--12020},
  year={2023}
}

@inproceedings{wang2023dreamwalker,
  title={Dreamwalker: Mental planning for continuous vision-language navigation},
  author={Wang, Hanqing and Liang, Wei and Van Gool, Luc and Wang, Wenguan},
  booktitle={Proceedings of the IEEE/CVF international conference on computer vision},
  pages={10873--10883},
  year={2023}
}

@inproceedings{krantz2021waypoint,
  title={Waypoint models for instruction-guided navigation in continuous environments},
  author={Krantz, Jacob and Gokaslan, Aaron and Batra, Dhruv and Lee, Stefan and Maksymets, Oleksandr},
  booktitle={Proceedings of the IEEE/CVF International Conference on Computer Vision},
  pages={15162--15171},
  year={2021}
}

@inproceedings{krantz2022sim,
  title={Sim-2-sim transfer for vision-and-language navigation in continuous environments},
  author={Krantz, Jacob and Lee, Stefan},
  booktitle={European conference on computer vision},
  pages={588--603},
  year={2022},
  organization={Springer}
}

@inproceedings{hong2022bridging,
  title={Bridging the gap between learning in discrete and continuous environments for vision-and-language navigation},
  author={Hong, Yicong and Wang, Zun and Wu, Qi and Gould, Stephen},
  booktitle={Proceedings of the IEEE/CVF conference on computer vision and pattern recognition},
  pages={15439--15449},
  year={2022}
}

@article{an2024etpnav,
  title={Etpnav: Evolving topological planning for vision-language navigation in continuous environments},
  author={An, Dong and Wang, Hanqing and Wang, Wenguan and Wang, Zun and Huang, Yan and He, Keji and Wang, Liang},
  journal={IEEE Transactions on Pattern Analysis and Machine Intelligence},
  year={2024},
  publisher={IEEE}
}

@inproceedings{raychaudhuri2021language,
  title={Language-aligned waypoint (law) supervision for vision-and-language navigation in continuous environments},
  author={Raychaudhuri, Sonia and Wani, Saim and Patel, Shivansh and Jain, Unnat and Chang, Angel},
  booktitle={Proceedings of the 2021 conference on empirical methods in natural language processing},
  pages={4018--4028},
  year={2021}
}

@inproceedings{georgakis2022cross,
  title={Cross-modal map learning for vision and language navigation},
  author={Georgakis, Georgios and Schmeckpeper, Karl and Wanchoo, Karan and Dan, Soham and Miltsakaki, Eleni and Roth, Dan and Daniilidis, Kostas},
  booktitle={Proceedings of the IEEE/CVF conference on computer vision and pattern recognition},
  pages={15460--15470},
  year={2022}
}

@article{chen2022weakly,
  title={Weakly-supervised multi-granularity map learning for vision-and-language navigation},
  author={Chen, Peihao and Ji, Dongyu and Lin, Kunyang and Zeng, Runhao and Li, Thomas and Tan, Mingkui and Gan, Chuang},
  journal={Advances in Neural Information Processing Systems},
  volume={35},
  pages={38149--38161},
  year={2022}
}

@inproceedings{chen2021topological,
  title={Topological planning with transformers for vision-and-language navigation},
  author={Chen, Kevin and Chen, Junshen K and Chuang, Jo and V{\'a}zquez, Marynel and Savarese, Silvio},
  booktitle={Proceedings of the IEEE/CVF Conference on Computer Vision and Pattern Recognition},
  pages={11276--11286},
  year={2021}
}

@inproceedings{savva2019habitat,
  title={Habitat: A platform for embodied ai research},
  author={Savva, Manolis and Kadian, Abhishek and Maksymets, Oleksandr and Zhao, Yili and Wijmans, Erik and Jain, Bhavana and Straub, Julian and Liu, Jia and Koltun, Vladlen and Malik, Jitendra and others},
  booktitle={Proceedings of the IEEE/CVF international conference on computer vision},
  pages={9339--9347},
  year={2019}
}

@inproceedings{krantz2020beyond,
  title={Beyond the nav-graph: Vision-and-language navigation in continuous environments},
  author={Krantz, Jacob and Wijmans, Erik and Majumdar, Arjun and Batra, Dhruv and Lee, Stefan},
  booktitle={European Conference on Computer Vision},
  pages={104--120},
  year={2020},
  organization={Springer}
}

@article{hong2020language,
  title={Language and visual entity relationship graph for agent navigation},
  author={Hong, Yicong and Rodriguez, Cristian and Qi, Yuankai and Wu, Qi and Gould, Stephen},
  journal={Advances in Neural Information Processing Systems},
  volume={33},
  pages={7685--7696},
  year={2020}
}

@inproceedings{fu2020counterfactual,
  title={Counterfactual vision-and-language navigation via adversarial path sampler},
  author={Fu, Tsu-Jui and Wang, Xin Eric and Peterson, Matthew F and Grafton, Scott T and Eckstein, Miguel P and Wang, William Yang},
  booktitle={European Conference on Computer Vision},
  pages={71--86},
  year={2020},
  organization={Springer}
}

@article{fried2018speaker,
  title={Speaker-follower models for vision-and-language navigation},
  author={Fried, Daniel and Hu, Ronghang and Cirik, Volkan and Rohrbach, Anna and Andreas, Jacob and Morency, Louis-Philippe and Berg-Kirkpatrick, Taylor and Saenko, Kate and Klein, Dan and Darrell, Trevor},
  journal={Advances in neural information processing systems},
  volume={31},
  year={2018}
}

@inproceedings{thomason2020vision,
  title={Vision-and-dialog navigation},
  author={Thomason, Jesse and Murray, Michael and Cakmak, Maya and Zettlemoyer, Luke},
  booktitle={Conference on Robot Learning},
  pages={394--406},
  year={2020},
  organization={PMLR}
}

@inproceedings{qi2020reverie,
  title={Reverie: Remote embodied visual referring expression in real indoor environments},
  author={Qi, Yuankai and Wu, Qi and Anderson, Peter and Wang, Xin and Wang, William Yang and Shen, Chunhua and Hengel, Anton van den},
  booktitle={Proceedings of the IEEE/CVF conference on computer vision and pattern recognition},
  pages={9982--9991},
  year={2020}
}

@inproceedings{ku2020room,
  title={Room-across-room: Multilingual vision-and-language navigation with dense spatiotemporal grounding},
  author={Ku, Alexander and Anderson, Peter and Patel, Roma and Ie, Eugene and Baldridge, Jason},
  booktitle={Proceedings of the 2020 Conference on Empirical Methods in Natural Language Processing (EMNLP)},
  pages={4392--4412},
  year={2020}
}

@inproceedings{anderson2018vision,
  title={Vision-and-language navigation: Interpreting visually-grounded navigation instructions in real environments},
  author={Anderson, Peter and Wu, Qi and Teney, Damien and Bruce, Jake and Johnson, Mark and S{\"u}nderhauf, Niko and Reid, Ian and Gould, Stephen and Van Den Hengel, Anton},
  booktitle={Proceedings of the IEEE conference on computer vision and pattern recognition},
  pages={3674--3683},
  year={2018}
}

@article{zhang2024uni,
  title={Uni-navid: A video-based vision-language-action model for unifying embodied navigation tasks},
  author={Zhang, Jiazhao and Wang, Kunyu and Wang, Shaoan and Li, Minghan and Liu, Haoran and Wei, Songlin and Wang, Zhongyuan and Zhang, Zhizheng and Wang, He},
  journal={arXiv preprint arXiv:2412.06224},
  year={2024}
}

@article{cheng2024navila,
  title={Navila: Legged robot vision-language-action model for navigation},
  author={Cheng, An-Chieh and Ji, Yandong and Yang, Zhaojing and Gongye, Zaitian and Zou, Xueyan and Kautz, Jan and B{\i}y{\i}k, Erdem and Yin, Hongxu and Liu, Sifei and Wang, Xiaolong},
  journal={arXiv preprint arXiv:2412.04453},
  year={2024}
}
